# Arabic natural language processing: An overview

Imane Guellil [a,*], Houda Saâdane [b], Faical Azouaou [a], Billel Gueni [c], Damien Nouvel [d]

[a] Laboratoire des Méthodes de Conception des Systèmes, Ecole nationale Supérieure d'Informatique, BP 68M, 16309 Oued-Smar, Alger, Algeria
[b] GEOLSemantics, 12 Avenue Raspail, 94250 Gentilly, France
[c] Responsable de la recherche et l'innovation Altran IT, Paris, France
[d] Inalco, France

## ABSTRACT

Arabic is recognised as the 4th most used language of the Internet. Arabic has three main varieties: (1) classical Arabic (CA), (2) Modern Standard Arabic (MSA), (3) Arabic Dialect (AD). MSA and AD could be written either in Arabic or in Roman script (Arabizi), which corresponds to Arabic written with Latin letters, numerals and punctuation. Due to the complexity of this language and the number of corresponding challenges for NLP, many surveys have been conducted, in order to synthesise the work done on Arabic. However these surveys principally focus on two varieties of Arabic (MSA and AD, written in Arabic letters only), they are slightly old (no such survey since 2015) and therefore do not cover recent resources and tools. To bridge the gap, we propose a survey focusing on 90 recent research papers (74% of which were published after 2015). Our study presents and classifies the work done on the three varieties of Arabic, by concentrating on both Arabic and Arabizi, and associates each work to its publicly available resources whenever available.

© 2019 The Authors. Production and hosting by Elsevier B.V. on behalf of King Saud University. This is an open access article under the CC BY-NC-ND license (http://creativecommons.org/licenses/by-nc-nd/4.0/).

## Contents

* Corresponding author.
  E-mail addresses: i_guellil@esi.dz (I. Guellil), f_azouaou@esi.dz (H. Saâdane), houda.saadane@geolsemantics.com (F. Azouaou), damien.nouvel@inalco.fr (D. Nouvel).
  URL: http://www.esi.dz (I. Guellil).
Peer review under responsibility of King Saud University.

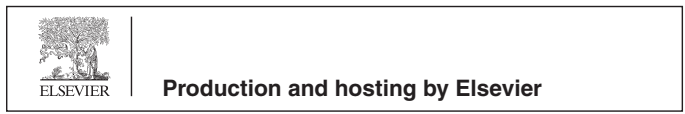

## 1. Introduction

Arabic is the official language of 22 countries, spoken by more than 400 million speakers. It is recognized as the 4th most used language of the Internet (Boudad et al., 2017). Arabic is classified in three main varieties (Habash, 2010; Farghaly and Shaalan, 2009; Harrat et al., 2017): 1) CA, a form of Arabic language used in literary texts and Quran[1] Sharaf and Atwell (2012a). 2) MSA, used for writing as well as formal conversations. 3) AD, used in daily life communication, informal exchanges, etc (Boudad et al., 2017). AD are mostly divided into six main groups: (1) Egyptian (EGY), (2) Levantine (LEV), (3) Gulf (GLF), (4) Iraqi (IRQ), (5) Maghrebi (MAGH) and (6) Others contains the remaining dialect (Habash, 2010; Sadat et al., 2014; Zaidan and Callison-Burch, 2014). However, Arabic speakers on social media, discussion forums and Short Messaging System (SMS) often use a non standard romanization called 'Arabizi' (Darwish, 2014; Bies et al., 2014). For example, the Arabic sentence: راني فرحانة, which means I am happy, is written in Arabizi as 'rani fer7ana'. Hence, Arabizi is an Arabic text written using Latin characters, numerals and some punctuation (Darwish, 2014).

Over the last decade, Arabic and its dialects have begun to gain ground in the area of research within Natural Language Processing (NLP). Much work targeted different aspects related to how this language and its dialects are processed, such as: Morphological analysis, resource building, Machine translation, etc. In order to present the characteristics of this language and to classify the works handling it, different surveys were proposed (Habash, 2010; Farghaly and Shaalan, 2009; Shoufan and Alameri, 2015). Habash gives a detailed overview of Arabic and its dialects by paying particular attention to different orthographic, morphological and syntactic aspects of this language (Habash, 2010). Farghaly and Shaalan (2009) described some challenges of Arabic NLP (ANLP) and presented solutions that would guide practitioners in this field. In the most recent ANLP survey of Shoufan and Alameri (2015), the authors proposed a general classification of achievements for Arabic and its dialects which can be grouped into four categories: (1) Basic Language Analyses (BLA). (2) Building Resources (BR), (3) Language Identification (LI) and (4) Semantic-Level Analysis (SemA). BLA concentrates on morphological, syntactic and orthographic analysis. The resources built could be lexicons corpora. LI is done on text and voice data. The tasks on which most work has been conducted in SemA are Machine Translation (MT) and Sentiment Analysis (SA). All the present surveys, in the literature, suffer from many drawbacks discussed in the following:

- All the surveys were presented before 2015, which is rather remote as a tremendous number of works has been done over the last three years.
- Arabic has three main varieties, CA, MSA and AD. However, MSA and AD could be written either in Arabic or in Roman script (Arabizi). Almost all the surveys presented in the literature principally focused on MSA and AD by neglecting the works done on CA. For Arabizi, the majority of works have been done after 2015, hence these surveys could not cover them.
- The research literature only presented the studies that have been done without presenting the resources and tools associated to the studies

To bridge the gap, this survey is dedicated to presenting and classifying the most recent works (90 studies) that have been done on Arabic. Most of them were published between 2015 and 2018 (exactly 67, which accounts for corresponding to 74% of the whole). It is the first survey, as far as we know, covering all varieties of Arabic such as: CA, MSA, DA by focusing on the two scripts used, Arabic and Arabizi. Another originality of this paper is the fact that a major part of the work (exactly 52 works representing 58% of the whole) are associated to freely and publicly available tools or resources.

This survey is organized as follows: Section 2 presents the work that have been done on CA. Section 3 presents the work on MSA. Section 4 presents the works on AD. Section 5 presents the works on Arabizi. Section 6 presents a synthesis of the studied works with a discussion on the most open issues in Arabic natural language processing. Finally, Section 7 presents the conclusion with some future directions.

## 2. Works on classical Arabic

Table 1 presents the studied works and projects focused on CA, especially on the Quran. More details about these works are given in the following sections.

### 2.1. Basic Language Analyses (BLA)

Dukes et al. presented an annotated linguistic resource (A part of Quranic Arabic Corpus noted QAC) The corpus provides three levels of analysis: morphological annotation, a syntactic treebank and a semantic ontology (Dukes et al., 2010). These authors introduced traditional Arabic grammar and describing the annotation process, including the syntactic relations used to label dependency graphs. They also highlight key parts of the full annotation guidelines such as: Verbs, Subjects and Objects and pronoun.

---

[1] The Quran is a scripture which, according to Muslims, is the verbatim words of Allah containing over 77,000 words (Sharaf and Atwell, 2012b).

**Table 1**
The studied works on Classical Arabic (Quran).

| Works and projects | Year | Research area | Resources types and link |
| --- | --- | --- | --- |
| Sharaf and Atwell (2012b) | 2012 | BR | Quranic corpus[1] Visualization tool [2] |
| Sharaf and Atwell (2012a) | 2012 | BR | Quranic corpus [3] Visualization tool [4] |
| Zerrouki and Balla (2017) | 2017 | BR | Vocalized[10] |
| Belinkov et al. (2016) | 2016 | BR | Shamela corpus |
| Dukes et al. (2010) | 2010 | BLA | Quranic corpus and Syntactic guideline [5] |
| Asda et al. (2016) | 2016 | LI | Identification system |
| Quran Analysis project | 2015 | SemA | Translation, transliteration and other semantic analysis[6] online tool[7] |
| Tanzil project | 2007 | SemA | Translation of Quran[8] Project program[9] |
| Al-Kabi et al. (2013) | 2013 | SemA | Classification system |
| Adeleke et al. (2017) | 2017 | SemA | Classification system |

[1] URL: http://textminingthequran.com/wiki/Verse_relatedness_in_Ibn_Kathir.
[2] URL: http://www.textminingthequran.com/apps/similarity.php.
[3] URL: http://corpus.quran.com/.
[4] URL: http://textminingthequran.com/wiki/Pronoun_Reference_in_the_Quran.
[5] URL: http://www.textminingthequran.com/apps/pron.php.
[6] URL: https://github.com/karimouda/qurananalysis.
[7] URL: http://www.qurananalysis.com/.
[8] URL: http://tanzil.net/trans/.
[9] URL: https://github.com/minijoomla/Tanzil.
[10] URL: https://sourceforge.net/projects/tashkeela/.

### 2.2. Building Resources (BR)

Sharaf and Atwell (2012b) presented a manually annotated large corpus (QurSim), created from the original Quranic text, where semantically similar or related verses are linked together. Sharaf and Atwell (2012a) also presents QurAna, a large corpus created from the original Quranic text, where personal pronouns are tagged with their antecedents. QurAna is characterized by a large number of pronouns (over 24,500 pronouns). Belinkov et al. (2016) proposed a large-scale historical Arabic corpus These authors lemmatized the entire corpus in order to use it in semantic analysis. This corpus contains 6,000 texts (1 billion words). Zerrouki and Balla (2017) propose a large freely available vocalized corpus, containing 75 million words, collected from freely published texts in old books.

### 2.3. Language Identification (LI)

Asda et al. (2016), propose the development of Quran reciter recognition and identification system, based on Mel-Frequency Cepstral Coefficient (MFCC) feature extraction and Artificial Neural Networks. From every speech, characteristics from the utterances will be extracted through neural network models. The proposed system is divided into two parts. The first part consists of feature extraction and the second part is the identification process using neural networks.

#### 2.3.1. Semantic-Level Analysis (SemA)

The team of Eric Atwell at Leeds University work on the Quran analysis project, in order to build a Semantic Search and Intelligence System. This system provides manual users with the ability to search the Quran semantically and analyze all aspects of the text. Another project dealing with a different task including translation is the Tanzil project. Tanzil is a Quranic project launched in early 2007 to produce a highly compliant unicode Quran text used in Quranic websites and applications. Among the applications of Tanzil features we cite the translation of Quran into other languages such as: English, German, Italian, etc. Al-Kabi et al. (2013) and Adeleke et al. (2017) focused on the classification of Quran. The first work aimed to classify different Quranic verses according to their topics and the second one presented a feature selection approach to automatically label Quranic verses.

## 3. Works on MSA

Table 2 presents works that have been conducted on MSA. Details on the presented works are given in the following sections.

### 3.1. Basic Language Analyses

Abdelali et al. presents a fast and accurate Arabic segmenter (Farasa) (Abdelali et al., 2016). This approach is based on SVM using linear kernels. To validate Farasa, the authors compare it (on two tasks MT and Information Retrieval IR) with two other segmenters: MADAMIRA (Pasha et al., 2014) and the Stanford Arabic Segmenter (SAS) (Monroe et al., 2014). Farassa outperforms both segmenter for IR task and it is at par with MADAMIRA for MT tasks. Howevers, MADAMIRA offers many tasks (such as: feature modelling, tokenization, phrase chunking and named entity recognition) which are not included in Farasa. Moreover, MADAMIRA is not dedicated to MSA only but also handle the Egyptian dialect. Abainia et al. (2017) presented ARLSTem, an Arabic light stemmer. The main idea of this stemmer is to remove the prefixes, suffixes and infixes (additional letter which is not original to the word root, present in the middle of the word).

Khalifa et al. (2016b) present YAMAMA, a morphological analyzer, focused on MSA and EGY dialect. YAMAMA was inspired by the fast execution of Farasa and the rich output of MADAMIRA. The results obtained with YAMAMA, in the context of MT, were better than those obtained with Farasa and similar to those obtained with MADAMIRA. More recently, Zalmout and Habash (2017) introduced a model for Arabic morphological disambiguation based on Recurrent Neural Networks (RNN). The authors based their work on Penn Arabic Treebank (PATB) (Maamouri et al., 2004) and they used Long Short-Term Memory model (LSTM) by showing that this model provides a significant performance.

Shahrour et al. (2016) presented CamelParser, a system for Arabic syntactic dependency analysis. CamelParser improves on the morphological disambiguation done in MADAMIRA (Pasha et al., 2014) using the syntactic analysis information. CamelParser improves the accuracy results of a former parser that the authors presented in Shahrour et al. (2015). More recently Taji et al. (2017) present NUDAR, a Universal Dependency treebank for Arabic. The authors presented the fully automated conversion from the Penn Arabic Treebank (PATB) (Maamouri et al., 2004) to the Universal Dependency syntactic representation. Arabic was also one of the studied languages on which More et al. (2018) applied their Universal Morphological Lattices for Universal Dependency (UD) Parsing. These authors presented CoNLL-UL annotation for morphological ambiguity, focuses on three language including Arabic.

**Table 2**
The studied works on Modern Standard Arabic (MSA).

| Works | Year | Research area | Resources types and link |
|---|---|---|---|
| Abdelali et al. (2016) | 2016 | BLA | Segmenter[1] |
| Pasha et al. (2014) | 2014 | BLA | MADAMIRA[2] |
| Abainia et al. (2017) | 2017 | BLA | ARLSTem[3] |
| Zalmout and Habash (2017) | 2017 | BLA | morphological disambiguation model |
| Khalifa et al. (2016b) | 2016 | BLA | YAMAMA[4] |
| Shahrour et al. (2015); Shahrour et al., 2016 | 2016 | BLA | Parser [5] |
| Taji et al. (2017) | 2016 | BLA | NUDAR[6] |
| More et al. (2018) | 2018 | BLA | CONLL-UL[7] |
| Selab and Guessoum (2015) | 2015 | BR + BLA | sample of corpus + code [8] |
| Yousfi et al. (2015) | 2015 | BR | corpus [9] |
| El-Haj and Koulali (2013) | 2013 | BR | KALIMAT corpus |
| Dima et al. (2018) | 2018 | BR | treebank dependency |
| Works | Year | Research area | Resources types and link |
| Ziemski et al. (2016) | 2016 | BR + MT | Parallel corpus UN (MSA + 5 others language)[10] |
| Lison and Tiedemann (2016) | 2016 | BR + MT | Parallel corpus (60 language including MSA)[11] |
| Inoue et al. (2018) | 2018 | BR + MT | Parallel corpus (MSA-Japanese)[12] |
| Badaro et al. (2014) | 2014 | BR + SA | Sentiment lexicon + [13] |
| Mohammad et al. (2016) | 2016 | BR + SA | Sentiment lexicon[14] |
| Al-Twairesh et al. (2016) | 2016 | BR + SA | Sentiment lexicon[15] |
| Rushdi-Saleh et al. (2011) | 2011 | BR + SA | Sentiment corpus[16] |
| Abdul-Mageed and Diab (2012) | 2012 | BR | Sentiment corpus |
| Aly and Atiya (2013) | 2013 | BR + SA | Sentiment corpus[17] |
| Dahou et al. (2016) | 2016 | SA | SA model |
| Badaro et al. (2018) | 2013 | EM | EM model |
| Kim et al. (2016) | 2016 | LM | LM model[18] |

[1] URL: http://qatsdemo.cloudapp.net/farasa/.
[2] URL: https://camel.abudhabi.nyu.edu/madamira/.
[3] URL: https://www.nltk.org/_modules/nltk/stem/arlstem.html.
[4] URL: https://nyuad.nyu.edu/en/research/centers-labs-and-projects/computational-approaches-to-modeling-language-lab/resources.html.
[5] URL: http://camel.abudhabi.nyu.edu/resources/.
[6] URL: https://github.com/UniversalDependencies/UD_Arabic-NYUAD.
[7] URL: https://conllul.github.io/.
[8] URL: https://github.com/saidziani/Arabic-News-Article-Classification.
[9] URL: https://cactus.orange-labs.fr/ALIF/.
[10] URL: https://conferences.unite.un.org/uncorpus.
[11] URL: http://opus.nlpl.eu/OpenSubtitles2016.php.
[12] URL: http://el.tufs.ac.jp/tufsmedia-corpus/.
[13] URL: http://www.oma-project.com/.
[14] URL: http://saifmohammad.com/WebPages/ArabicSA.html.
[15] URL: https://github.com/nora-twairesh/AraSenti.
[16] URL: http://sinai.ujaen.es/en/?s=oca&submit=Search.
[17] URL: https://github.com/mahmoudnabil/labr.
[18] URL: https://github.com/yoonkim/lstm-char-cnn.

### 3.2. Building Resources

Selab and Guessoum (2015) construct the TALAA corpus, a large Arabic corpus (containing 14 million words), built from daily Arabic newspaper websites (57,827 different articles). Yousfi et al. (2015) present the dataset ALIF dedicated to Arabic embedded text recognition in TV broadcasts. It is composed of a large number of manually annotated text images that were extracted from Arabic TV broadcasts. The ALIF corpus contains 6,532 Arabic text color images. El-Haj and Koulali (2013) presented KALIMAT corpus (20,291 Arabic articles), collected from the Omani newspaper Alwatan The authors proceed to a set of actions on the collected data such as: (1) Summarization , (2) Recognition of the named entities and (3) Annotation of the data collection. Dima et al. (2018) presented a small dependency treebank of travel domain sentences in Modern Standard Arabic. The corpus is created by translating the selected 2,000 sentences from the Basic Travelling Expression Corpus (BTEC) presented by Takezawa in Takezawa (2006).

Different parallel corpora were proposed for MSA such as those presented by: 1) Ziemski et al. (2016) dealing, in addition to MSA, with five other languages: Chinese, English, French, Russian, Spanish. This corpus contains 799,276 documents and 1,727,539 aligned document pairs. 2) Inoue et al. (2018) for MSA and Japanese. A part of this corpus comprising 8,652 document was manually aligned at the sentence level for development and testing and 3) Lison and Tiedemann (2016) which proposes a subtitled parallel corpus covering 60 languages including Arabic. This corpus is the release of an extended and improved version of the OpenSubtitles[2] collection of parallel corpora.

Different sentiment lexicons (Badaro et al., 2014; Mohammad et al., 2016; Al-Twairesh et al., 2016) and corpora (Rushdi-Saleh et al., 2011; Abdul-Mageed and Diab, 2012; Aly and Atiya, 2013) were constructed in order to perform sentiment analysis tasks. Badaro et al. (2014) proposed ArSenL, a large-scale sentiment lexicon (157,969 synonymous and 28,760 lemmas) by exploiting available Arabic and English resources. Following the example of ArSenL (Badaro et al., 2014), Eskander and Rambow (2015)presented SLSA (34,821 entries), constructed by linking the lexicon of an Arabic morphological analyzer with SentiWordNet (Baccianella et al., 2010). Mohammad et al. (2016) used distant supervision techniques and they translated an existing English sentiment lexicons into Arabic using Google Translate. Rushdi-Saleh et al. (2011) presented OCA (an Opinion Corpus for Arabic), a manual annotated corpus, containing 500 movie reviews (250 positive and 250 negative) collected from different Arabic web pages and blogs. Abdul-Mageed and Diab (2012) presented AWATIF, a diversified corpus containing 10,723 Arabic sentences, manually annotated as objective or subjective (positives and negatives). Aly and Atiya (2013) present LABR containing 63,257 book reviews rated on a scale from 1 to 5 stars. The authors considered reviews with 4 or 5 stars as positive, those with 1 or 2 stars as negative, and ones with 3 stars as neutral.

---

[2] http://www.opensubtitles.org.

### 3.3. Identification and recognition

Many works are oriented on Arabic identification (Ali et al., 2016; El Haj et al., 2017; Shon et al., 2018; Tachicart et al., 2017). These works were respectively presented by Ali et., El Haj et al., Shon et al. and Tachicart et al. They dealt with the Multi dialectal identification as well as the MSA one where the last one proposes an identification system distinguishing between the Moroccan dialect and MSA. However, as the main focus of these works is related to dialect identification, they are presented in more detail in Section 4.3.

### 3.4. Semantic-level analysis and synthesis

Almost all of the MT system are based on statistical machine translation (using Moses toolkit (Koehn et al., 2007)) which requires a parallel corpus. In this context, Inoue et al. (2018) present the first results of Arabic-Japanese phrase-based MT by relying on the alignment of 900 documents using two techniques: manual and automatic. Their system returned a BLEU score up to 9.38. Ziemski et al. (2016), focus only on 4,000 sentences that they have randomly taken from the UN corpus. The results for the pair Arabic/English were up to 53.07 and those for the pair English/Arabic are up to 41.96. For aligning sentences, Lison and Tiedemann (2016) relied on the time-overlap algorithm employed in Tiedemann (2007). The BLEU score for the pair Arabic/English was up to 25.34.

In the context of SA system, Badaro et al. (2014) presents a non-linear SVM implementation in MATLAB, with the radial basis function (RBF) kernel The best F1-score achieved by the authors is up to 64.5%. To determine the usefulness of their Arabic sentiment lexicons, Mohammad et al. (2016) applied them in a sentence-level sentiment analysis system. The authors used a linear-kernel SVM (Chang and Lin, 2011) classifier. The best F1-score was up to 66.6%. Al-Twairesh et al. (2016) performed a set of experiments using a simple lexicon-based approach. Their results show the importance on handling negation and they achieved an F1-score up to 89.58% Rushdi-Saleh et al. (2011) have used cross-validation to compare the performance of SVM and NB algorithms. The best accuracy was achieved with SVM classifier and it is up to 0.91. Aly and Atiya (2013) used Multinomial NB, Bernoulli NB and SVM. The best F1-score (up to 0.91) was achieved with the unbalanced dataset combining unigrams, bigrams and trigrams, by using TF-IDF and with SVM classifiers. Dahou et al. (2016) introduce a method based on word2vec which evaluates polarity from product reviews. A convolutional neural network (CNN) model was trained on a model of Arabic word embeddings. More recently, Badaro et al. (2018) proposed EMA system (Emotion Mining in Arabic) The authors proceed to a set of treatments including: Diacritics, normalisation, stemming (using ARLSTEM (Abainia et al., 2017)), etc. To extract features, the authors used the two embedding vectors AraVec (Soliman et al., 2017) and fasText (Joulin et al., 2016). The authors used SVM, Random Forest (RF) and a set of deep neural network algorithms such as CNN with LSTM LSTM layer. The last work (Kim et al., 2016) focus on proposing a language model. In this context, Kim et al. (2016) propose a language model based on character-level CNN. The output of this model is used as an input to a recurrent neural network language model (RNN-LM). The authors applied their approach on 7 languages including Arabic where they used the OPUS corpus.[3]

## 4. Works on Arabic dialects

The following table summarizes many works related to Arabic dialects. These works will be detailed in the next sections.

### 4.1. Basic Language Analyses

Many works have been proposed in order to offer a set of Orthographic rules, standards and conventions. The work presented by Saadane and Habash (2015) follows the previous efforts made and demonstrated for Egyptian and Tunisian dialects and applies them for the Algerian dialect. The purpose of Habash et al. (2018) was to present a common set of guidelines with enough specificity to help in creating dialect specific conventions. Segmentation and Part-of-speech (POS) tagging are two of the most important addressed areas NLP. More attention has been given recently to process Arabic Dialects (Samih et al., 2017; Alharbi et al., 2018; Darwish et al., 2018). Samih et al. (2017) present a segmenter which is trained on 350 annotated tweets using neural networks. In their model, the authors consider Arabic segmentation as a character-based sequence classification problem. Alharbi et al. (2018) present a recent POS taggers for the Arabic Gulf dialect. The authors observed that using a Bi-LSTM labeler consequently improved the results obtained with SVM. Darwish et al. (2018) also present a POS tagger which relies on a Conditional Random Fields (CRF) sequence labeler. This tagger is dedicated to the four important Dialect classes such as: EGY, LEV, GLF, and MAGH. To validate their approach the authors have manually segmented a set of 350 tweets in each dialect.

Salloum and Habash (2014) presented ADAM (Analyzer for Dialectal Arabic Morphology). The authors evaluated ADAM's performance on LEV and EGY. ADAM is comparable in its performance to CALIMA (Habash et al., 2012), which is an EGY dialectal morphological analyzer that required years and expensive resources to build. In the same way, Khalifa et al. (2017) presents a GLF Arabic morphological analyzer covering over 2,600 verbs. The authors employed two resources (a collection of root-abstracted paradigms and a lexicon of verbs) Zribi et al. (2013) proposed a method adapting a MSA morphological analyzer for the Tunisian dialect (TD). To do that, they relied on TD lexicons that they constructed based on an existing MSA lexicon. Al-Shargi et al. (2016) presented a morphological analyzer and tagger concentrating on the Morrocan and Sanaani Yemeni dialects. This analyzer was trained on a morphologically annotated corpus that the authors constructed manually using the annotation interface DIWAN (Al-Shargi and Rambow, 2015). Khalifa et al. (2018) also presented a large-scale morphologically manually annotated corpus of Emirati Arabic. They relied on 200,000 words selected from Gumar (Khalifa et al., 2016a). For the annotation, the authors used MADARi interface (Obeid et al., 2018). MADARi is a web-based interface supporting both morphological annotation and spelling correction during all the process of annotation. MADARi was initially using MADAMIRA (for EGY dialect). However, it was extended with CALIMAGLF for more coverage. Zalmout et al. (2018) recently presented a neural morphological tagging and disambiguation model for the Egyptian dialect, with various extensions to handle noisy and inconsistent content. these authors relied on LSTM and CNN model for generating their character embedding.

---

[3] http://opus.nlpl.eu/News-Commentary.php.

*4.2. Building Resources*

Kwaik et al. (2018) present the construction of the Shami corpus, a LEV Dialect Corpus. This corpus covers data from the four dialects spoken in Palestine, Jordan, Lebanon and Syria and contains 117,805 sentences. Jarrar et al. (2017) present the construction of *Curras*, a morphologically annotated corpus of the Palestinian Arabic dialect. *Curras* consists of more than 56,000 tokens, which were annotated with rich morphological and lexical features. Al-Twairesh et al. (2018) proposed SUAR, a semi-automatically Saudi corpus which was morphologically annotated automatically using the MADAMIRA (Pasha et al., 2014). The generated corpus was manually checked The resulted corpus contains 104,079 words. Alsarsour et al. (2018) presented DART (The Dialectal ARabic Tweets dataset), a large manually-annotated multi-dialect corpus of Arabic tweets This corpus contains around 25,000 tweets annotated via crowd-sourcing. Erdmann et al. (2018) proposed different word embedding models for the four dialects EGY, LEV, MAGH and GLF. To build a corpus as large as possible, the authors combined different earlier corpora (Almeman and Lee, 2013; Khalifa et al., 2016a; Bouamor et al., 2018). The resulting corpus contains 5.6 million sentences Suwaileh et al. (Suwaileh et al., 2016) presented ArabicWeb16, a multi-domain, very large corpus containing 10.8 TB.[4] Khalifa et al. (2016a) focus on GLF Dialects where they proposed a large-scale corpus consisting of 110 million words from 1,200 novels. This corpus contains many domains such as; romance, drama and tragedy. More recently Abdul-Mageed et al. (2018) built a large-scale dataset for a variety of Arabic dialects. This corpus contains 234,801,907 tweets.

Saad and Alijla (2017) propose the construction of a comparable Wikipedia corpus between the Arabic and Egyptian dialects. This corpus contains 10,197 aligned documents Another comparable corpus (CALYOU) was constructed by Abidi et al. (2017). This corpus is dedicated to spoken Algerian To align messages, the authors used different approaches such as: dictionary based, indexing words by their sounds and finally an approach based on the similarity proposed by word2vec (Mikolov et al., 2013). Bouamor et al. (2018) presented the construction of two valuable resources: (1) a parallel corpus of 25 Arabic dialects. (2) A lexicon containing 47,466 dialectal words covering the 25 studied Arabic dialects. The corpus is created by translating (BTEC) corpus (Takezawa, 2006) Tachicart et al. (2014) concentrate on the presentation of an Arabic dialect lexicon (bilingual lexicon between MSA and the Moroccan dialect) by translating an MSA dictionary.[5] After some eliminations and manual validation, the authors obtained 18,000 entries in their lexicon. Kumar et al. (2014) presented (Callhome), an EGY Arabic-English Speech Translation corpus. The speech part of the corpus consists of telephone conversations between native speakers. Each conversation is about 5–30 min; Gender, age, education and accent of speakers were also added.

El-Beltagy (2016) presents NileULex (containg 5,953 unique terms), an Arabic sentiment lexicon containing EGY and MSA. Abdul-Mageed and Diab (2016) presents SANA (224,564 entries), a large-scale manually annotated multi-genre, multi-dialectal and multi-lingual lexical resource for subjectivity and sentiment analysis of Arabic and dialects. In addition to MSA, SANA covers also both EGY and LEV, and also providing English glosses. Guellil et al. (2017d) presented an Algerian sentiment lexicon constructed automatically by relying on two English lexicon (SOCAL (Taboada et al., 2011) and SentiWordNet (Esuli and Sebastiani, 2007)) that they automatically translated using glosbeAPI.[6] The resulted lexicons contains. 3952 entries for SentiAlg and 2375 for SOCALAlg. Medhaffar et al. (2017) presented the TSAC (Tunisian Sentiment Analysis Corpus) corpus containing 17,060 Tunisian Facebook comments, manually annotated (8215 positive and 8845 negative). Guellil et al. (2018b) presented the automatic construction of an Algerian sentiment corpus using the constructed lexicon (Guellil et al., 2017d) Afterwards, they randomly selected 8000 messages (where 4000 are positives and 4000 are negatives) Al-Twairesh et al. (2017), introduced the corpus AraSenTi-Tweet containing 17,573 Saudi tweets and semi-automatically annotated into four classes: positive, negative, neutral and mixed. To construct this corpus, the authors target a set of sentiment words and use them to extract tweets containing these words. Afterwards, they manually reviewed the automatically constructed corpus.

*4.3. Language Identification*

Ali et al. (2016), used different approaches for dialect identification in Arabic broadcast speech focused on multi dialects (EGY, LEV, GLF, and MAGH as well as MSA). Their methods are based on phonetic and lexical features obtained from a speech recognition system El Haj et al. (2017) present an approach of Arabic dialect identification using language bivalency[7] and written code-switching. The authors concentrate on multi dialects as well as MSA. For the classification task, the authors use different algorithms: NB, SVM, etc. More recently Ali (2018) proposed a character-level convolution neural network model for distinguishing between MSA and multi dialects. The authors proposed a CNN model including five layers Shon et al. (2018) proposed an end-to-end Dialect identification system and a Siamese neural network to extract language embeddings. The authors used acoustic and linguistic features on the Arabic dialectal speech dataset. The authors use the MGB-3 (Multi-Genre Broadcast) dataset (Ali et al., 2017). Tachicart et al. (2017) focused on an Identification system distinguishing between the Moroccan Dialect and MSA. The authors relied on two different approaches: (1) rule-based and (2) statistical-based (using several machine learning classifiers). However the statistical approach outperforms the rule-based approach where the SVM classifier is more accurate than other statistical classifiers. The work of Salameh and Bouamor (2018) bring a new idea by presenting the first results on a fine-grained dialect classification covering 25 specific Arabic cities (such as Morocco, Egypt, Iraq, Gulf, etc), in addition to MSA. To do so, the authors relied on MADAR, the parallel corpus of 25 Arabic dialects (Bouamor et al., 2018). The authors consider the identification task as a classification task where they used a Multinomial NB (MNB) classifier for the learning task.

*4.4. Semantic-level analysis*

In the context of MT, Meftouh et al. (2018) used a phrase-based MT system, GIZA++ (Och and Ney, 2003) for alignment and SRILM toolkit (Stolcke, 2002) The best results that these authors obtained were between the Algiers dialect and the dialect of Annaba (with BLEU score up to 67.31) which is perfectly understandable where both dialects are spoken into the same country (Algeria). To the best of our knowledge, Saad and Alijla (2017), Abidi et al. (2017), Bouamor et al. (2018), Kumar et al. (2014) have not proposed any system to validate their comparable and parallel corpora.

In the context of SA, El-Beltagy (2016) implement a simple sentiment analysis task using the bag of words model, with uni-gram and bi-gram TF-IDF weights. As a classifier, the authors used NB in combination with the constructed lexicon (NileULex). The results show that the integration of NileULex improved the results of classification (F1-score up to 79%). Guellil et al. (2017d) propose a sentiment analysis algorithm dealing with Algerian dialect morphology and handling negation and opposition. The best F1-score that they achieved is up to 78%. To evaluate the performance of their corpus, Medhaffar et al. (2017) relied on three classifiers such as SVM and NB and MultiLayer Perceptron classifier (MLP). The input of each sentiment classifier is the set of feature vectors from the Doc2vec (Le and Mikolov, 2014) toolkit. The authors showed that SVM and MLP gave better results while poorer results are obtained by using the BN classifier. Guellil et al. (2018b), relied on two vectorization techniques which are: Bag Of Words (BOW) and Doc2vec. For classification, they used different classifiers such as SVM, NB, etc. Al-Twairesh et al. (2017) conducted several experiments for multi-way sentiment classification. For classification, they used SVM with a linear kernel. For the term feature, they tested the term-presence, term-frequency, and TF-IDF features. The best results were achieved with two-way classification and term Presence feature.

---

[4] 1 TeraByte (TB) = one trillion bytes.
[5] https://ia802304.us.archive.org/27/items/xvmf6/xvmf6.pdf.
[6] https://glosbe.com/.

[7] A word or element is treated by language users as having a fundamentally similar semantic content in more than one language or dialect (El Haj et al., 2017).

## 5. Works on Arabizi

### 5.1. Basic Language Analyses

Guellil and Azouaou (2016b) proposed ASDA, a Syntactic Analyzer for Algerian dialect(final state automaton). This parser extracted from the same term several parts (corresponding to verbs, nouns, adjectives, conjunctions, and the different pronouns). These authors focus on Algerian Arabizi. To build this analyzer, the authors initially enrich a basic dictionary that contains translation between the Algerian dialect and French, with different phonological extensions.

### 5.2. Building Resources

The idea of the work Guellil and Faical (2017); Azouaou and Guellil, 2017 is to present a bilingual lexicon between Algerian Arabizi and French. The authors first relied on an existing resource which they pre-processed and also enriched. To enrich their lexicon, the authors principally worked on the phonological aspects of Arabizi. They proposed to extend each word by all the different words that are pronounced in the same manner. The resulted lexicon contains 25,086 words

### 5.3. Arabizi identification

Darwish (2014) classified a word as Arabizi or English in-context, by using a sequence labeling based on Conditional Random Fields (CRF). This system achieved an accuracy of 98.5% for word-level language identification. Guellil and Azouaou (2016a) proposed an approach for Arabic dialect identification in social media based on supervised methods (using a pre-built lexicon proposed previously (Guellil and Faical, 2017; Azouaou and Guellil, 2017)). The primary goal of this approach is to identify the words written in the Algerian Arabizi dialect. The authors applied their approach on 100 messages manually annotated and they achieved an accuracy up to 60%.

### 5.4. Semantic-level analysis

In the context of MT, Guellil et al. (2017a); Guellil et al., 2017b presented a translation system between the Algerian Arabizi and MSA. The authors proposed a comparison between statistical and neural translations after the transliteration step. These authors showed that the quality of transliteration directly affects the translation, where BLEU score was up to 4.26 for direct translation (i.e without transliteration) and up to 6.01 (for automatic transliteration) and up to 10.74 (for manual transliteration). May et al. (2014) used a phrase-based SMT system similar to Moses (Koehn et al., 2007), trained on the collection of Arabic-English parallel corpora The translation results between Arabic and English are up to 9.89 (BLEU score, after manual transliteration). van der Wees et al. (2016) built an SMT system. They began with the unsupervised alignment of words in hand-aligned sentences. The best results achieved by this approach are up to 18.4 (manual transliteration) and up to 15.1 (automatic transliteration).

In the context of SA, Duwairi et al. (2016) constructed a small annotated corpus manually (containing 3,026 messages) containing Arabizi messages transliterated into Arabic. For classification, these authors used NB and SVM classifiers where SVM gave better results. Some other works try dealing with Arabizi without the transliteration step (Medhaffar et al., 2017; Guellil et al., 2018b). However, Guellil et al. (2018b) clearly stated that the low F1-score that they obtained for the Arabizi dataset (up to 0.66) is principally related to the fact of handling Arabizi without transliteration. In addition to Arabizi, the two works of Medhaffar et al. (2017) and Guellil et al. (2018b) also concentrated on Arabic, they are presented in detail in Section 4.3.

### 5.5. Arabizi transliteration

Three Arabizi transliteration approach are emerging: 1) Rule-based approaches (Habash et al., 2007; Eskander et al., 2014; Guellil et al., 2018a; Duwairi et al., 2016). 2) Statistical approaches (Rosca and Breuel, 2016) and 3) Hybrid approaches. (Al-Badrashiny et al., 2014; Guellil et al., 2017c; Guellil et al., 2017a; Guellil et al., 2017b; Darwish, 2014; May et al., 2014; van der Wees et al., 2016). Rosca and Breuel (2016) proposed a statistical approach by defining a model based on neural networks (proposed in Graves et al. (2006); Bahdanau et al., 2014) to perform transliteration between several language pairs including Arabic and English. The work proposed by Al-Badrashiny et al. (2014), Guellil et al. (2017c), Guellil et al. (2017a), Guellil et al. (2017b), Darwish (2014), May et al. (2014), van der Wees et al. (2016), Guellil et al. (2018) share the same general idea. These authors firstly generate or extract a set of candidates for transliteration (the possible transliterations) and then, they determine the best candidate using a language model or other.

## 6. Synthesis and discussion

A total of 90 research papers focusing on ANLP were presented and classified. Our first purpose was to focus on the most recent research work; in this respect 67 papers (74.4%) of the studied works were published between 2015 and 2018. The presented works are done on CA (10 works corresponding to 11.1%), MSA (24 works, 26.7%), AD (30 works, 33.3%), Arabizi (14 works, 15.6%). Some works also combine MSA/AD (10 works, 11.1%) and AD/Arabizi (2 works, 2.2%). We were also able to identify 52 resources and tools (57.8%). However, it can be seen from Table 4, that no Arabizi work could be associated to its resource or tool (which represent 17.8%). Concerning the work on CA, MSA and AD, we were able to associate 52/74 works (70.3%).

From the aforementioned statistics, we observe that the majority of work have been done on Arabic dialects rather than CA, MSA or Arabizi. However, Arabizi represents an emerging research area where 12 (which represents 75%) works among the 16 presented are between 2015 and 2018. It can be seen from Table 3 that an

**Table 3**
The studied works on Dialectal Arabic (DA).

| Works | Year | Research area | Dialect type | Resources types and link |
|---|---|---|---|---|
| Darwish et al. (2018) | 2018 | BLA | Multi | Tagger (POS)[1] |
| Saadane and Habash (2015) | 2015 | BLA | Algerian | CODA Guideline |
| Habash et al. (2018) | 2012 | BLA | Multi | Orthography Guideline[2] |
| Alharbi et al. (2018) | 2018 | BLA | Gulf | Tagger (POS)[3] and [4] |
| Samih et al. (2017) | 2017 | BLA | Egyptian | Tagger (POS)[5] |
| Salloum and Habash (2014) | 2014 | BLA | Levantine + Egyptian | ADAM[6] |
| Khalifa et al. (2017) | 2017 | BLA | Gulf | $CALIMA_{GLF}$[7] |
| Zribi et al. (2013) | 2013 | BLA | Tunisian | Corpus[8] |
| Al-Shargi et al. (2016) | 2016 | BLA | Morrocan + Sanaani Yemeni | morphological analyzer |
| Al-Shargi and Rambow (2015) | 2015 | BLA | Multi | morphological annotation |
| Khalifa et al. (2018) | 2018 | BLA + BR | Emirati (Gulf) | morphological, pos annotation[9] |
| Obeid et al. (2018) | 2018 | BLA | Egyptian + Gulf | MADARi [10] |
| Zalmout et al. (2018) | 2018 | BLA | Egyptian | morphological disambiguation |
| Ali et al. (2016) | 2016 | LI | Multi + MSA | Identification[11] |
| El Haj et al. (2017) | 2017 | LI | Multi + MSA | Identification[12] +corpus[13] |
| Ali (2018) | 2018 | LI | Multi- MSA | Identification[14] |
| Shon et al. (2018) | 2018 | LI | Multi + MSA | Identification[15] |
| Tachicart et al. (2017) | 2017 | LI | Moroccan + MSA | Identification[16] |
| Salameh and Bouamor (2018) | 2018 | LI | Multi | Identification |
| Maamouri et al. (2014) | 2014 | BR + BLA | Egyptian | Egyptian Treebank + morphological analyser |
| Kwaik et al. (2018) | 2018 | BR | Levantine | Corpus + Identification[17] |
| Al-Twairesh et al. (2018) | 2018 | BR | Saudi | Saudi corpus |
| Meftouh et al. (2018) | 2018 | BR + MT | Multi | PADIC[18] |
| Saad and Alijla (2017) | 2017 | BR | EGY-MSA | Comparable corpus[19] + WikiDocsAligner[20] |
| Bouamor et al. (2018) | 2018 | BR | Multi (25 dialects) | Madar parallel corpus and lexicon[21] |
| Alsarsour et al. (2018) | 2018 | BR | Multi | DART corpus[22] |
| Erdmann et al. (2018) | 2018 | BR | Multi | Corpus[23] |
| Tachicart et al. (2014) | 2014 | BR | Moroccan-MSA | Lexicon[24] |
| Abidi et al. (2017) | 2017 | BR | Algerian | Comparable corpus |
| Jarrar et al. (2017) | 2017 | BR | Palestinian | Corpus[25] |
| Suwaileh et al. (2016) | 2016 | BR | Multi-MSA | Corpus[26] |
| Khalifa et al. (2016a) | 2016 | BR | Gulf-MSA | Online interface[27] |
| Abdul-Mageed et al. (2018) | 2018 | BR | Multi | Extraction tweet code[28] |
| Kumar et al. (2014) | 2014 | BR | EGY-English | Parallel corpus[29] |
| El-Beltagy (2016) | 2016 | BR + SA | EGY-MSA | Sentiment lexicon[30] |
| Abdul-Mageed and Diab (2016) | 2014 | BR + SA | Multi | Sentiment lexicon |
| Medhaffar et al. (2017) | 2017 | BR + SA | Tunisian | Sentiment corpus[31] |
| Guellil et al. (2018b) | 2018 | BR + SA | Algerian | Sentiment corpus |
| Al-Twairesh et al. (2017) | 2017 | BR + SA | Saudi | Sentiment corpus + SA |
| Guellil et al. (2017d) | 2017 | BR + SA | Algerian | Sentiment lexicon + SA |

[1] URL: https://github.com/qcri/dialectal_arabic_resources.
[2] URL: http://resources.camel-lab.com/.
[3] URL: https://github.com/qcri/dialectal_arabic_pos_tagger.
[4] URL: http://alt.qcri.org/resources/da_resources/.
[5] URL: https://github.com/qcri/dialectal_arabic_tools.
[6] URL: https://github.com/WaelSalloum/adam.
[7] URL: http://camel.abudhabi.nyu.edu/resources/.
[8] URL: https://github.com/NadiaBMKarmani/Intelligent-Tunisian-Arabic-Morphological-Analyzer-evaluation-corpus.
[9] URL: https://camel.abudhabi.nyu.edu/gumar/.
[10] URL: https://nyuad.nyu.edu/en/research/centers-labs-and-projects/computational-approaches-to-modeling-language-lab/resources.html.
[11] URL: https://github.com/qcri/dialectID.
[12] URL: https://github.com/drelhaj/ArabicDialects.
[13] URL: http://www.lancaster.ac.uk/staff/elhaj/corpora.htm.
[14] URL: https://github.com/bigoooh/adi.
[15] URL: https://github.com/swshon/dialectID_e2e.
[16] URL: http://arabic.emi.ac.ma:8080/MCAP/faces/lid.xhtml;jsessionid=6c824eea3d4d42560be0a8e429b0.
[17] URL: https://github.com/GU-CLASP/shami-corpus.
[18] URL: http://smart.loria.fr/pmwiki/pmwiki.php/PmWiki/Corpora.
[19] URL: https://github.com/motazsaad/comparableWikiCorpus.
[20] URL: https://github.com/motazsaad/WikiDocsAligner.
[21] URL: http://nlp.qatar.cmu.edu/madar/.
[22] URL: http://qufaculty.qu.edu.qa/telsayed/datasets/.
[23] URL: https://camel.abudhabi.nyu.edu/arabic-multidialectal-embeddings/.
[24] URL: http://arabic.emi.ac.ma:8080/mded/#MdedListForm:j_idt20:j_idt21.
[25] URL: http://portal.sina.birzeit.edu/curras/download.html.
[26] URL: http://qufaculty.qu.edu.qa/telsayed/arabicweb16.
[27] URL: http://camel.abudhabi.nyu.edu/gumar/.
[28] URL: https://github.com/hasanhuz/Location_Analysis_Project.
[29] URL: https://github.com/noisychannel/ARZ_callhome_corpus.
[30] URL: https://github.com/NileTMRG/NileULex.
[31] URL: https://github.com/fbougares/TSAC.

Table 4
The studied works on Arabizi

| Works | Year | Research area | Arabizi type | Resources types and link |
|---|---|---|---|---|
| Guellil and Azouaou (2016b) | 2017 | BLA | Algerian | Parser |
| Guellil and Faical (2017) | 2017 | BR | Algerian/ French | Bilingual lexicon |
| Azouaou and Guellil (2017) | 2017 | BR | Algerian/ French | Bilingual lexicon |
| Darwish (2014) | 2014 | BR + LI + TR | MSA + Multi | corpus Arabizi-MSA + identification + TR |
| Guellil and Azouaou (2016a) | 2016 | LI | Algerian | Identification |
| Guellil et al. (2017a) | 2017 | TR + BR + MT | Algerian | Corpus Arabizi-MSA + TR + MT |
| Guellil et al. (2017b) | 2017 | TR + BR + MT | Algerian | Corpus Arabizi-MSA + TR + MT system |
| Eskander et al. (2014) | 2014 | TR + Classification | Egyptian | Arabizi transliteration + classification system |
| May et al. (2014) | 2017 | TR + BR + MT | MSA + Multi | Corpus Arabizi-MSA + TR + MT |
| van der Wees et al. (2016) | 2016 | TR + BR + MT | MSA + Multi | Corpus Arabizi-MSA + TR + MT |
| Duwairi et al. (2016) | 2016 | TR + BR + SA | MSA + Multi | Corpus Arabizi-MSA + TR + SA |
| Habash et al. (2007) | 2007 | TR | MSA + Multi | TR system |
| Guellil et al. (2018) | 2018 | TR | Algerian | TR system |
| Guellil et al. (2017c) | 2017 | TR + BR | Algerian | Corpus Arabizi-MSA + TR |
| Al-Badrashiny et al. (2014) | 2014 | TR | MSA + Multi | TR system |
| Guellil et al. (2018a) | 2018 | TR + BR + SA | Algerian | SA corpus + TR/SA system |

important number of works are focused on multi-dialects (exactly 17/40, which represents 42.5% of the dialect works). For the rest of work on dialects (i.e. 23/40 which represents 57.5%), they mostly focused on Saudi/GLF (i.e. 7/23) and EGY (6/23). The work on MAGH dialects are less sizeable particularly for Tunisian and Moroccan (i.e 2/23 for each one) and for Algerian (4/23). The situation is worse for the Palestinian and LeV dialects (with 1/23 work for each one). We also observe that the resources and tools are rarely publicly available for some dialects. For example, for the Algerian dialect, only one resource (PADIC) is publicly available. However, resources and tools are more abundant in other dialects such as GLF and EGY.

From Table 4, we observe two trends related to Arabizi works: 1) work handling Arabizi in general (as one Arabic language) and 2) work considering Arabizi as any possible dialect (focusing on Algerian Arabizi or Egyptian Arabizi). However, the most recent work tends to follow the second trend. This is understandable, since Arabizi is only a Latin transliteration of Arabic and its dialects. Unfortunately, no work deals with Arabizi identification for distinguishing between Algerian, Tunisian, Egyptian Arabizi, etc. Some work deals with identification but it deals with the identification between Arabizi, French and English. It does not handle the intra dialect identification for Arabizi.

Finally, we also observe that most studied works concentrated on building resources (lexicon and corpora). Even if the main purpose of these works is SemA, BLA, LI or Tr, resource construction is crucial. Mainly due to this reason, 50 works among the 90 provide resources (which represent 55.6%). However almost of the presented resources were constructed manually which is time and effort-consuming. By considering Arabic and its dialects as an under-resourced language, almost all the recent studies are working on proposing new resources for bridging the gap. The most studied research area is BLA (with 24 works out of 90, 26.7%). This is mainly due to the morphological complexity of Arabic and its dialects which makes this language difficult to handle without pre-processing illustrated by POS, Stemming, lemmatisation, etc.

## 7. Conclusion and perspectives

The presented paper contains analysis and classification of 90 recent research work and project covering all Arabic varieties, written with both script (i.e Arabic and Arabizi). The originality of this survey is to associate works to their publicly available tools and resources. From the analysis of the presented works, we conclude that only few works have been done on CA. Hence, this Arabic variety contains many open issues. Arabizi represents an emerging research area but all the presented resources are not yet publicly available. Although almost all works have targeted MSA and AD, these two varieties are still considered under-resources languages compared to more studied languages such as English.

This survey opens the door to many research questions: (1) Is it better to rely on manually built resources or to propose methods and techniques to create such resources automatically? (2) Is it really crucial to transliterate before any semantic analysis? (3) Is it better to handle each dialect individually, or is it possible to propose methods and techniques able to process all dialects? (4) Should we always associate AD to MSA or could it be better to associate AD to other languages such as English, French, etc? (5) Why do research studies always follow the way of construction? Is it possible to rely on existing resources, to combine resources, etc? (6) Are deep learning approaches really more efficient than traditional approaches such as SVM, NB, etc, for Arabic natural processing?

The main idea of this survey is to present to the community of research the most recent resources and tools. In the near future, we plan to propose another survey paper where we use the presented publicly available tools and resources and where we apply them to given extrinsic data, for presenting to the community the strength and weakness of each resources.

## References

Abainia, K., Ouamour, S., Sayoud, H., 2017. A novel robust arabic light stemmer. J. Exp. Theor. Artif. Intell. 29, 557–573.

Abdelali, A., Darwish, K., Durrani, N., Mubarak, H., 2016. Farasa: a fast and furious segmenter for arabic. In: Proceedings of the 2016 Conference of the North American Chapter of the Association for Computational Linguistics: Demonstrations, pp. 11–16.

Abdul-Mageed, M., Alhuzali, H., Elaraby, M., 2018. You tweet what you speak: a city-level dataset of arabic dialects. LREC.

Abdul-Mageed, M., Diab, M.T., 2012. Awatif: A Multi-genre Corpus for Modern Standard Arabic Subjectivity and Sentiment Analysis. LREC, Citeseer, pp. 3907–3914.

Abdul-Mageed, M., Diab, M.T., 2016. Sana: A Large Scale Multi-genre, Multi-dialect Lexicon for Arabic Subjectivity and Sentiment Analysis. LREC.

Abidi, K., Menacer, M.A., Smaili, K., 2017. Calyou: A comparable spoken algerian corpus harvested from youtube. In: 18th Annual Conference of the International Communication Association (Interspeech).

Adeleke, A.O., Samsudin, N.A., Mustapha, A., Nawi, N.M., 2017. Comparative analysis of text classification algorithms for automated labelling of quranic verses. Int. J. Adv. Sci. Eng. Info. Tech. 7, 1419–1427.

Al-Badrashiny, M., Eskander, R., Habash, N., Rambow, O., 2014. Automatic transliteration of romanized dialectal arabic. In: Proceedings of the Eighteenth Conference on Computational Natural Language Learning, pp. 30–38.

Al-Kabi, M.N., Alsmadi, I.M., Wahsheh, H.A., Abu Ata, B.M., 2013. A topical classification of quranic arabic text, NOORIC 2013: Taibah University International Conference on Advances in Information Technology for the Holy Quran and Its Sciences.

Al-Shargi, F., Kaplan, A., Eskander, R., Habash, N., Rambow, O., 2016. Morphologically annotated corpora and morphological analyzers for moroccan and sanaani yemeni arabic. In: 10th Language Resources and Evaluation Conference (LREC 2016).

Al-Shargi, F., Rambow, O., 2015. Diwan: A dialectal word annotation tool for arabic. In: Proceedings of the Second Workshop on Arabic Natural Language Processing, pp. 49–58.

Al-Twairesh, N., Al-Khalifa, H., Al-Salman, A., Al-Ohali, Y., 2017. Arasenti-tweet: a corpus for arabic sentiment analysis of saudi tweets. Proc. Comput. Sci. 117, 63–72.

Al-Twairesh, N., Al-Khalifa, H., AlSalman, A., 2016. Arasenti: large-scale twitter-specific arabic sentiment lexicons. In: Proceedings of the 54th Annual Meeting of the Association for Computational Linguistics (volume 1: Long Papers), pp. 697–705.

Al-Twairesh, N., Al-Matham, R., Madi, N., Almugren, N., Al-Aljmi, A.H., Alshalan, S., Alshalan, R., Alrumayyan, N., Al-Manea, S., Bawazeer, S., et al., 2018. Suar: Towards building a corpus for the Saudi dialect. Proc. Comput. Sci. 142, 72–82.

Alharbi, R., Magdy, W., Darwish, K., AbdelAli, A., Mubarak, H., 2018. Part-of-speech tagging for arabic gulf dialect using bi-lstm.

Ali, A., Dehak, N., Cardinal, P., Khurana, S., Yella, S.H., Glass, J., Bell, P., Renals, S., 2016. Automatic dialect detection in arabic broadcast speech. In: Interspeech, San Francisco, CA, USA, pp. 2934–2938.

Ali, A., Vogel, S., Renals, S., 2017. Speech recognition challenge in the wild: Arabic mgb-3. In: Automatic Speech Recognition and Understanding Workshop (ASRU), 2017 IEEE. IEEE, pp. 316–322.

Ali, M., 2018. Character level convolutional neural network for arabic dialect identification. In: Proceedings of the Fifth Workshop on NLP for Similar Languages, Varieties and Dialects (VarDial 2018), pp. 122–127.

Almeman, K., Lee, M., 2013. Automatic building of arabic multi dialect text corpora by bootstrapping dialect words. In: Communications, signal processing, and their applications (iccspa), 2013 1st international conference on, Citeseer, pp. 1–6.

Alsarsour, I., Mohamed, E., Suwaileh, R., Elsayed, T., 2018. Dart: a large dataset of dialectal arabic tweets. In: Proceedings of the Eleventh International Conference on Language Resources and Evaluation (LREC-2018).

Aly, M., Atiya, A., 2013. Labr: a large scale arabic book reviews dataset. In: Proceedings of the 51st Annual Meeting of the Association for Computational Linguistics, pp. 494–498 (Volume 2: Short Papers).

Asda, T.M.H., Gunawan, T.S., Kartiwi, M., Mansor, H., 2016. Development of quran reciter identification system using mfcc and neural network.

Azouaou, F., Guellil, I., 2017. Alg/fr: A step by step construction of a lexicon between algerian dialect and french, in: The 31st Pacific Asia Conference on Language, Information and Computation PACLIC 31 (2017).

Baccianella, S., Esuli, A., Sebastiani, F., 2010. Sentiwordnet 3.0: an enhanced lexical resource for sentiment analysis and opinion mining. LREC.

Badaro, G., Baly, R., Hajj, H., Habash, N., El-Hajj, W., 2014. A large scale arabic sentiment lexicon for arabic opinion mining. In: Proceedings of the EMNLP 2014 Workshop on Arabic Natural Language Processing (ANLP), pp. 165–173.

Badaro, G., El Jundi, O., Khaddaj, A., Maarouf, A., Kain, R., Hajj, H., El-Hajj, W., 2018. Ema at semeval-2018 task 1: Emotion mining for arabic. In: Proceedings of The 12th International Workshop on Semantic Evaluation, pp. 236–244.

Bahdanau, D., Cho, K., Bengio, Y., 2014. Neural machine translation by jointly learning to align and translate. arXiv preprint arXiv:1409.0473.

Belinkov, Y., Magidow, A., Romanov, M., Shmidman, A., Koppel, M., 2016. Shamela: a large-scale historical arabic corpus. arXiv preprint arXiv:1612.08989.

Bies, A., Song, Z., Maamouri, M., Grimes, S., Lee, H., Wright, J., Strassel, S., Habash, N., Eskander, R., Rambow, O., 2014. Transliteration of arabizi into arabic orthography: developing a parallel annotated arabizi-arabic script sms/chat corpus. In: Proceedings of the EMNLP 2014 Workshop on Arabic Natural Language Processing (ANLP), pp. 93–103.

Bouamor, H., Habash, N., Salameh, M., Zaghouani, W., Rambow, O., Abdulrahim, D., Obeid, O., Khalifa, S., Eryani, F., Erdmann, A., et al., 2018. The madar arabic dialect corpus and lexicon. LREC.

Boudad, N., Faizi, R., Thami, R.O.H., Chiheb, R., 2017. Sentiment analysis in arabic: a review of the literature. Ain Shams Eng. J.

Chang, C.C., Lin, C.J., 2011. Libsvm: a library for support vector machines. ACM Trans. Intell. Syst. Technol. (TIST) 2, 27.

Dahou, A., Xiong, S., Zhou, J., Haddoud, M.H., Duan, P., 2016. Word embeddings and convolutional neural network for arabic sentiment classification. In: Proceedings of COLING 2016, the 26th International Conference on Computational Linguistics: Technical Papers, pp. 2418–2427.

Darwish, K., 2014. Arabizi detection and conversion to arabic. In: Proceedings of the EMNLP 2014 Workshop on Arabic Natural Language Processing (ANLP), pp. 217–224.

Darwish, K., Mubarak, H., Abdelali, A., Eldesouki, M., Samih, Y., Alharbi, R., Attia, M., Magdy, W., Kallmeyer, L., 2018. Multi-dialect Arabic Pos Tagging: A CRF Approach. LREC.

Dima, T., Jamila, E.G., Nizar, H., 2018. An arabic dependency treebank in the travel domain. In: 11th Language Resources and Evaluation Conference.

Dukes, K., Atwell, E., Sharaf, A.B.M., 2010. Syntactic Annotation Guidelines for the Quranic Arabic Dependency Treebank. LREC.

Duwairi, R.M., Alfaqeh, M., Wardat, M., Alrabadi, A., 2016. Sentiment analysis for arabizi text, in: Information and Communication Systems (ICICS). 2016 7th International Conference on, IEEE. pp. 127–132.

El-Beltagy, S.R., 2016. Nileulex: A Phrase and Word Level Sentiment Lexicon for Egyptian and Modern Standard Arabic. LREC.

El-Haj, M., Koulali, R., 2013. Kalimat a multipurpose arabic corpus. In: Second Workshop on Arabic Corpus Linguistics (WACL-2), pp. 22–25.

El Haj, M., Rayson, P.E., Aboelezz, M., 2017. Arabic dialect identification in the context of bivalency and code-switching.

Erdmann, A., Zalmout, N., Habash, N., 2018. Addressing noise in multidialectal word embeddings. In: Proceedings of the 56th Annual Meeting of the Association for Computational Linguistics, pp. 558–565 (Volume 2: Short Papers).

Eskander, R., Al-Badrashiny, M., Habash, N., Rambow, O., 2014. Foreign words and the automatic processing of arabic social media text written in roman script. Proceedings of The First Workshop on Computational Approaches to Code Switching, 1–12.

Eskander, R., Rambow, O., 2015. Slsa: A sentiment lexicon for standard arabic. In: Proceedings of the 2015 Conference on Empirical Methods in Natural Language Processing, pp. 2545–2550.

Esuli, A., Sebastiani, F., 2007. Sentiwordnet: a high-coverage lexical resource for opinion mining. Evaluation 17, 1–26.

Farghaly, A., Shaalan, K., 2009. Arabic natural language processing: challenges and solutions. ACM Transactions on Asian Language Information Processing (TALIP) 8, 14.

Graves, A., Fernández, S., Gomez, F., Schmidhuber, J., 2006. Connectionist temporal classification: labelling unsegmented sequence data with recurrent neural networks. In: Proceedings of the 23rd international conference on Machine learning ACM, pp. 369–376.

Guellil, I., Adeel, A., Azouaou, F., Benali, F., Hachani, A.E., Hussain, A., 2018. Arabizi sentiment analysis based on transliteration and automatic corpus annotation. In: Proceedings of the 9th Workshop on Computational Approaches to Subjectivity, Sentiment and Social Media Analysis, pp. 335–341.

Guellil, I., Adeel, A., Azouaou, F., Hussain, A., 2018b. Sentialg: Automated corpus annotation for algerian sentiment analysis. In: 9th International Conference on Brain Inspired Cognitive Systems(BICS 2018).

Guellil, I., Azouaou, F., 2016. Arabic dialect identification with an unsupervised learning (based on a lexicon). application case: Algerian dialect. In: Computational Science and Engineering (CSE) and IEEE Intl Conference on Embedded and Ubiquitous Computing (EUC) and 15th Intl Symposium on Distributed Computing and Applications for Business Engineering (DCABES), 2016 IEEE Intl Conference on, IEEE. pp. 724–731.

Guellil, I., Azouaou, F., 2016. Asda: Analyseur syntaxique du dialecte algérien dans un but d'analyse sémantique. In: Conférence Nationale d'Intelligence Artificielle Année 2016, Association Française pour l'Intelligence Artificielle AFIA. afia.asso. fr/wp. pp. 87–94.

Guellil, I., Azouaou, F., Abbas, M., 2017. Comparison between neural and sta-tistical translation after transliteration of algerian arabic dialect. In: WiNLP: Women & Underrepresented Minorities in Natural Language Processing (co-located withACL 2017).

Guellil, I., Azouaou, F., Abbas, M., 2017. Neural vs statistical translation of algerian arabic dialect written with arabizi and arabic letter. In: The 31st Pacific Asia Conference on Language, Information and Computation PACLIC 31 (2017).

Guellil, I., Azouaou, F., Abbas, M., Fatiha, S., 2017. Arabizi transliteration of algerian arabic dialect into modern standard arabic. In: Social MT 2017: First workshop on Social Media and User Generated Content Machine Translation (co-located with EAMT 2017).

Guellil, I., Azouaou, F., Benali, F., Hachani, A.E., Saadane, H., 2018. Approche hybride pour la translitération de l'arabizi algérien: une étude préliminaire. In: Conference: 25e conférence sur le Traitement Automatique des Langues Naturelles (TALN), May 2018, Rennes, FranceAt: Rennes, France,https://www.researchgate.net/publication.

Guellil, I., Azouaou, F., Saâdane, H., Semmar, N., 2017. Une approche fondée sur les lexiques d'analyse de sentiments du dialecte algérien.

Guellil, I., Faical, A., 2017. Bilingual lexicon for algerian arabic dialect treatment in social media. In: WiNLP: Women & Underrepresented Minorities in Natural Language Processing (co-located with ACL 2017).http://www.winlp.org/wp-content/uploads/2017/final_papers_2017/92_Paper.pdf.

Habash, N., Eryani, F., Khalifa, S., Rambow, O., Abdulrahim, D., Erdmann, A., Faraj, R., Zaghouani, W., Bouamor, H., Zalmout, N., et al., 2018. Unified Guidelines and Resources for Arabic Dialect Orthography. LREC.

Habash, N., Eskander, R., Hawwari, A., 2012. A morphological analyzer for egyptian arabic. In: Proceedings of the twelfth meeting of the special interest group on computational morphology and phonology. Association for Computational Linguistics, pp. 1–9.

Habash, N., Soudi, A., Buckwalter, T., 2007. On arabic transliteration. In: Arabic Computational Morphology. Springer, pp. 15–22.

Habash, N.Y., 2010. Introduction to arabic natural language processing. Synthesis Lectures on Human Language Technologies 3, 1–187.

Harrat, S., Meftouh, K., Smaïli, K., 2017. Maghrebi arabic dialect processing: an overview. In: ICNLSSP 2017-International Conference on Natural Language, Signal and Speech Processing.

Inoue, G., Habash, N., Matsumoto, Y., Aoyama, H., 2018. A Parallel Corpus of Arabic-japanese News Articles. LREC.

Jarrar, M., Habash, N., Alrimawi, F., Akra, D., Zalmout, N., 2017. Curras: an annotated corpus for the palestinian arabic dialect. Lang. Resour. Eval. 51, 745–775.

Joulin, A., Grave, E., Bojanowski, P., Mikolov, T., 2016. Bag of tricks for efficient text classification. arXiv preprint arXiv:1607.01759.

Khalifa, S., Habash, N., Abdulrahim, D., Hassan, S., 2016. A large scale corpus of gulf arabic. arXiv preprint arXiv:1609.02960.

Khalifa, S., Habash, N., Eryani, F., Obeid, O., Abdulrahim, D., Al Kaabi, M., 2018. A morphologically annotated corpus of emirati arabic. In: Proceedings of the Eleventh International Conference on Language Resources and Evaluation (LREC-2018).

Khalifa, S., Hassan, S., Habash, N., 2017. A morphological analyzer for gulf arabic verbs. In: Proceedings of the Third Arabic Natural Language Processing Workshop, pp. 35–45.

Khalifa, S., Zalmout, N., Habash, N., 2016b. Yamama: Yet another multi-dialect arabic morphological analyzer. In: Proceedings of COLING 2016, the 26th International Conference on Computational Linguistics: System Demonstrations, pp. 223–227.

Kim, Y., Jernite, Y., Sontag, D., Rush, A.M., 2016. Character-aware neural language models.

Koehn, P., Hoang, H., Birch, A., Callison-Burch, C., Federico, M., Bertoldi, N., Cowan, B., Shen, W., Moran, C., Zens, R., et al., 2007. Moses: Open source toolkit for statistical machine translation. In: Proceedings of the 45th annual meeting of the ACL on interactive poster and demonstration sessions. Association for Computational Linguistics, pp. 177–180.

Kumar, G., Cao, Y., Cotterell, R., Callison-Burch, C., Povey, D., Khudanpur, S., 2014. Translations of the Callhome Egyptian Arabic Corpus for Conversational Speech Translation. IWSLT, Citeseer.

Kwaik, K.A., Saad, M., Chatzikyriakidis, S., Dobnik, S., 2018. Shami: A corpus of levantine arabic dialects.

Le, Q., Mikolov, T., 2014. Distributed representations of sentences and documents. International Conference on Machine Learning, 1188–1196.

Lison, P., Tiedemann, J., 2016. Opensubtitles 2016: Extracting large parallel corpora from movie and tv subtitles.

Maamouri, M., Bies, A., Buckwalter, T., Mekki, W., 2004. The penn arabic treebank: building a large-scale annotated arabic corpus. In: NEMLAR Conference on Arabic Language Resources and Tools, Cairo, pp. 466–467.

Maamouri, M., Bies, A., Kulick, S., Ciul, M., Habash, N., Eskander, R., 2014. Developing an egyptian arabic treebank: impact of dialectal morphology on annotation and tool development. LREC.

May, J., Benjira, Y., Echihabi, A., 2014. An arabizi-english social media statistical machine translation system. In: Proceedings of the 11th Conference of the Association for Machine Translation in the Americas, pp. 329–341.

Medhaffar, S., Bougares, F., Esteve, Y., Hadrich-Belguith, L., 2017. Sentiment analysis of tunisian dialects: Linguistic resources and experiments. In: Proceedings of the Third Arabic Natural Language Processing Workshop, pp. 55–61.

Meftouh, K., Harrat, S., Smaïli, K., 2018. Padic: extension and new experiments. In: 7th International Conference on Advanced Technologies. ICAT.

Mikolov, T., Sutskever, I., Chen, K., Corrado, G.S., Dean, J., 2013. Distributed representations of words and phrases and their compositionality. Adv. Neural Inf. Processing Systems, 3111–3119.

Mohammad, S., Salameh, M., Kiritchenko, S., 2016. Sentiment Lexicons for Arabic Social Media. LREC.

Monroe, W., Green, S., Manning, C.D., 2014. Word segmentation of informal arabic with domain adaptation. In: Proceedings of the 52nd Annual Meeting of the Association for Computational Linguistics, pp. 206–211 (Volume 2: Short Papers).

More, A., Çetinoğlu, Ö., Çöltekin, Ç., Habash, N., Sagot, B., Seddah, D., Taji, D., Tsarfaty, R., 2018. Conll-ul: Universal morphological lattices for universal dependency parsing. In: 11th Language Resources and Evaluation Conference.

Obeid, O., Khalifa, S., Habash, N., Bouamor, H., Zaghouani, W., Oflazer, K., 2018. Madari: A web interface for joint arabic morphological annotation and spelling correction. arXiv preprint arXiv:1808.08392.

Och, F.J., Ney, H., 2003. A systematic comparison of various statistical alignment models. Comput. Ling. 29, 19–51.

Pasha, A., Al-Badrashiny, M., Diab, M.T., El Kholy, A., Eskander, R., Habash, N., Pooleery, M., Rambow, O., Roth, R., 2014. Madamira: A Fast, Comprehensive Tool for Morphological Analysis and Disambiguation of Arabic. LREC, pp. 1094–1101.

Rosca, M., Breuel, T., 2016. Sequence-to-sequence neural network models for transliteration. arXiv preprint arXiv:1610.09565.

Rushdi-Saleh, M., Martín-Valdivia, M.T., Ureña-López, L.A., Perea-Ortega, J.M., 2011. Oca: opinion corpus for arabic. J. Assoc. Inf. Sci. Technol. 62, 2045–2054.

Saad, M., Alijla, B.O., 2017. Wikidocsaligner: an off-the-shelf wikipedia documents alignment tool. In: Information and Communication Technology (PICICT), 2017 Palestinian International Conference on. IEEE, pp. 34–39.

Saadane, H., Habash, N., 2015. A conventional orthography for algerian arabic. In: ANLP Workshop 2015, p. 69.

Sadat, F., Mallek, F., Boudabous, M., Sellami, R., Farzindar, A., 2014. Collaboratively constructed linguistic resources for language variants and their exploitation in NLP application – the case of tunisian arabic and the social media. Proceedings of Workshop on Lexical and Grammatical Resources for Language Processing, 102–110.

Salameh, M., Bouamor, H., 2018. Fine-grained arabic dialect identification. In: Proceedings of the 27th International Conference on Computational Linguistics, pp. 1332–1344.

Salloum, W., Habash, N., 2014. Adam: analyzer for dialectal arabic morphology. J. King Saud Univ.-Comput. Inf. Sci. 26, 372–378.

Samih, Y., Attia, M., Eldesouki, M., Abdelali, A., Mubarak, H., Kallmeyer, L., Darwish, K., 2017. A neural architecture for dialectal arabic segmentation. In: Proceedings of the Third Arabic Natural Language Processing Workshop, pp. 46–54.

Selab, E., Guessoum, A., 2015. Building talaa, a free general and categorized arabic corpus. in: ICAART (1), pp. 284–291.

Shahrour, A., Khalifa, S., Habash, N., 2015. Improving arabic diacritization through syntactic analysis. In: Proceedings of the 2015 Conference on Empirical Methods in Natural Language Processing, pp. 1309–1315.

Shahrour, A., Khalifa, S., Taji, D., Habash, N., 2016. Camelparser: a system for arabic syntactic analysis and morphological disambiguation. In: Proceedings of COLING 2016, the 26th International Conference on Computational Linguistics: System Demonstrations, pp. 228–232.

Sharaf, A.B.M., Atwell, E., 2012a. Qurana: Corpus of the Quran Annotated with Pronominal Anaphora. LREC, Citeseer.

Sharaf, A.B.M., Atwell, E., 2012b. Qursim: A Corpus for Evaluation of Relatedness in Short Texts. LREC, Citeseer.

Shon, S., Ali, A., Glass, J., 2018. Convolutional neural networks and language embeddings for end-to-end dialect recognition. arXiv preprint arXiv:1803.04567.

Shoufan, A., Alameri, S., 2015. Natural language processing for dialectical arabic: a survey. In: Proceedings of the Second Workshop on Arabic Natural Language Processing, pp. 36–48.

Soliman, A.B., Eissa, K., El-Beltagy, S.R., 2017. Aravec: A set of arabic word embedding models for use in arabic NLP. Proc. Comput. Sci. 117, 256–265.

Stolcke, A., 2002. Srilm-an extensible language modeling toolkit. In: Seventh International Conference on Spoken Language processing.

Suwaileh, R., Kutlu, M., Fathima, N., Elsayed, T., Lease, M., 2016. Arabicweb16: A new crawl for today's arabic web. In: Proceedings of the 39th International ACM SIGIR conference on Research and Development in Information Retrieval ACM, pp. 673–676.

Taboada, M., Brooke, J., Tofiloski, M., Voll, K., Stede, M., 2011. Lexicon-based methods for sentiment analysis. Comput. Linguistics 37, 267–307.

Tachicart, R., Bouzoubaa, K., Aouragh, S.L., Jaafa, H., 2017. Automatic identification of moroccan colloquial arabic. In: International Conference on Arabic Language Processing. Springer, pp. 201–214.

Tachicart, R., Bouzoubaa, K., Jaafar, H., 2014. Building a moroccan dialect electronic dictionary (MDED). In: 5th International Conference on Arabic Language Processing, pp. 216–221.

Taji, D., Habash, N., Zeman, D., 2017. Universal dependencies for arabic. In: Proceedings of the Third Arabic Natural Language Processing Workshop, pp. 166–176.

Takezawa, T., 2006. Multilingual spoken language corpus development for communication research. In: Chinese Spoken Language Processing. Springer, pp. 781–791.

Tiedemann, J., 2007. Improved sentence alignment for movie subtitles. In: Proceedings of RANLP.

van der Wees, M., Bisazza, A., Monz, C., 2016. A simple but effective approach to improve arabizi-to-english statistical machine translation. In: Proceedings of the 2nd Workshop on Noisy User-generated Text (WNUT), pp. 43–50.

Yousfi, S., Berrani, S.A., Garcia, C., 2015. Alif: A dataset for arabic embedded text recognition in tv broadcast. In: Document Analysis and Recognition (ICDAR), 2015 13th International Conference on, IEEE. pp. 1221–1225.

Zaidan, O.F., Callison-Burch, C., 2014. Arabic dialect identification. Comput. Linguistics 40, 171–202.

Zalmout, N., Erdmann, A., Habash, N., 2018. Noise-robust morphological disambiguation for dialectal arabic. In: Proceedings of the 2018 Conference of the North American Chapter of the Association for Computational Linguistics: Human Language Technologies, Volume 1 (Long Papers), pp. 953–964.

Zalmout, N., Habash, N., 2017. Don't throw those morphological analyzers away just yet: neural morphological disambiguation for arabic, in. In: Proceedings of the 2017 Conference on Empirical Methods in Natural Language Processing, pp. 704–713.

Zerrouki, T., Balla, A., 2017. Tashkeela: novel corpus of arabic vocalized texts, data for auto-diacritization systems. Data in Brief 11, 147.

Ziemski, M., Junczys-Dowmunt, M., Pouliquen, B., 2016. The United Nations Parallel Corpus v1. LREC.

Zribi, I., Khemakhem, M.E., Belguith, L.H., 2013. Morphological analysis of tunisian dialect. In: Proceedings of the Sixth International Joint Conference on Natural Language Processing, pp. 992–996.